\documentclass[letterpaper, 10 pt, conference]{ieeeconf}  % Comment this line out if you need a4paper

\IEEEoverridecommandlockouts                              % This command is only needed if 
\usepackage{amsmath} % assumes amsmath package installed
\usepackage{amssymb}  % assumes amsmath package installed
\usepackage[usenames, dvipsnames]{color}
\usepackage[short]{optidef}
\usepackage{mathtools}
\usepackage{algorithm}
\usepackage{algorithmicx}
\usepackage{pgfplots}
\usepackage{graphicx}
\usepackage[abs]{overpic}
\usepackage{booktabs}
\usepackage{caption}
\usepackage{algpseudocode}
\usepackage{hyperref}
\usepackage{xcolor}
\usepackage{cuted}
\usepackage[export]{adjustbox}
\usepackage{float}
\usepackage{stfloats}
\usepackage{needspace}

\usetikzlibrary{arrows.meta}
\usetikzlibrary{backgrounds}
\usepgfplotslibrary{patchplots}
\usepgfplotslibrary{fillbetween}
\pgfplotsset{%
    layers/standard/.define layer set={%
        background,axis background,axis grid,axis ticks,axis lines,axis tick labels,pre main,main,axis descriptions,axis foreground%
    }{
        grid style={/pgfplots/on layer=axis grid},%
        tick style={/pgfplots/on layer=axis ticks},%
        axis line style={/pgfplots/on layer=axis lines},%
        label style={/pgfplots/on layer=axis descriptions},%
        legend style={/pgfplots/on layer=axis descriptions},%
        title style={/pgfplots/on layer=axis descriptions},%
        colorbar style={/pgfplots/on layer=axis descriptions},%
        ticklabel style={/pgfplots/on layer=axis tick labels},%
        axis background@ style={/pgfplots/on layer=axis background},%
        3d box foreground style={/pgfplots/on layer=axis foreground},%
    },
}

\algdef{SE}[DOWHILE]{Do}{doWhile}{\algorithmicdo}[1]{\algorithmicwhile\ #1}%
\usepackage[style=ieee,doi=false,isbn=false,url=false,eprint=false]{biblatex}
\title{\LARGE \bf
The Surprising Effectiveness of Linear Models for Whole-Body Model-Predictive Control
}

\author{
  Arun L. Bishop$^1$, Juan Alvarez-Padilla$^1$, Sam Schoedel$^1$, Ibrahima Sory Sow$^1$, 
Juee Chandrachud$^1$,\\ Sheitej Sharma$^1$, Will Kraus$^1$, Beomyeong Park$^2$, 
Robert J. Griffin$^2$, John M. Dolan$^1$, Zachary Manchester$^1$% <-this % stops a space
\thanks{$^1$Authors are with the Robotics Institute, the Department of Electrical and Computer Engineering, and the Department of
Mechanical Engineering, Carnegie Mellon University}% <-this % stops a space
\thanks{$^2$Authors are with the Florida Institute for Human and Machine Cognition}
}%

\begin{document}

\makeatletter
\let\@oldmaketitle\@maketitle% Store \@maketitle
\renewcommand{\@maketitle}{%
  \@oldmaketitle% Update
  \centering
  \includegraphics[width=\linewidth]{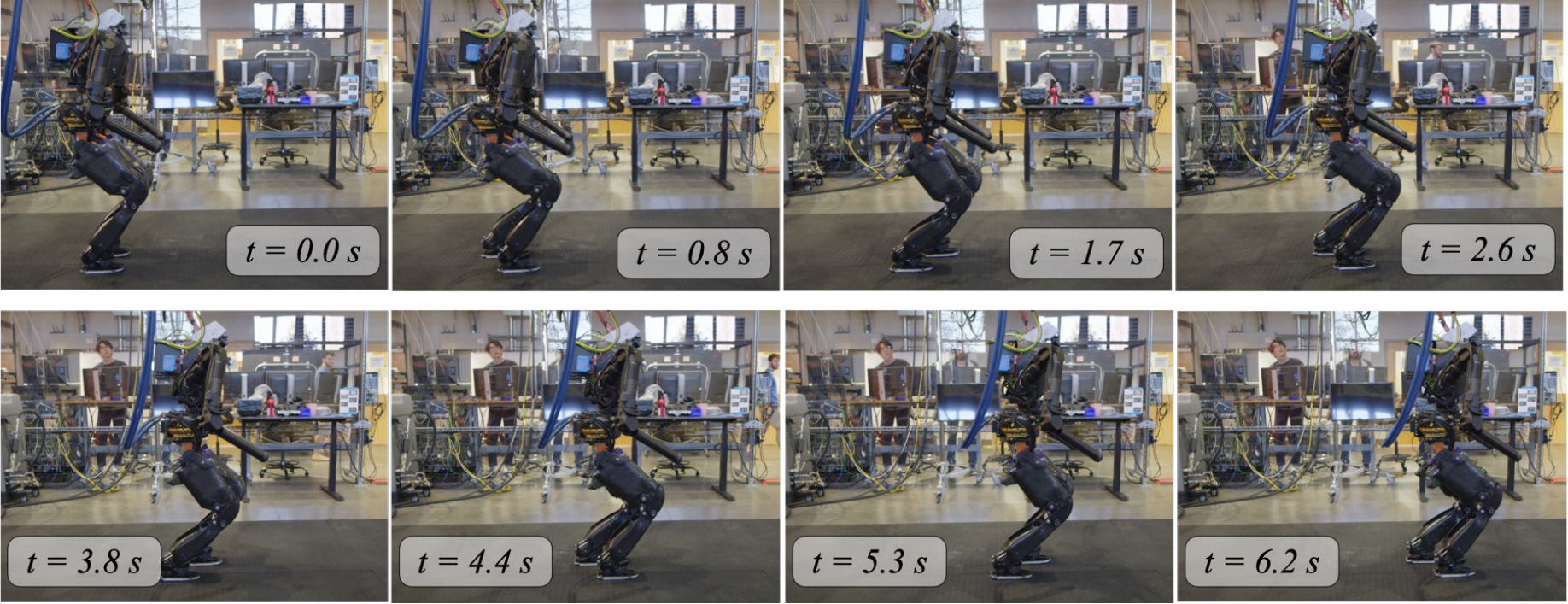}
  \captionof{figure}{
    The hydraulic humanoid Nadia walking forward over multiple steps using our convex MPC formulation with linear time-invariant dynamics constraints. The gait has a stride length of 30 cm, a swing phase of 0.6 s and a double support phase of 0.3 s.
  }
  \label{fig:nadia_walk}
  \vspace{-0.12in}
}
\makeatother

\let\oldmaketitle\maketitle
\renewcommand{\maketitle}{%
  \oldmaketitle
  \addtocounter{figure}{-1}  % Subtract one from figure counter
}

\maketitle

\thispagestyle{empty}
\pagestyle{empty}

%%%%%%%%%%%%%%%%%%%%%%%%%%%%%%%%%%%%%%%%%%%%%%%%%%%%%%%%%%%%%%%%%%%%%%%%%%%%%%%%

%%%%%%%%%%%%%%%%%%%%%%%%%%%%%%%%%%%%%%%%%%%%%%%%%%%%%%%%%%%%%%%%%%%%%%%%%%%%%%%%

\begin{abstract}
When do locomotion controllers require reasoning about nonlinearities? In this work, we show that a whole-body model-predictive controller using a simple linear time-invariant approximation of the whole-body dynamics is able to execute basic locomotion tasks on complex legged robots. The formulation requires no online nonlinear dynamics evaluations or matrix inversions. We demonstrate walking, disturbance rejection, and even navigation to a goal position without a separate footstep planner on a quadrupedal robot. In addition, we demonstrate dynamic walking on a hydraulic humanoid, a robot with significant limb inertia, complex actuator dynamics, and large sim-to-real gap.
\end{abstract}

\section{Introduction}

Legged locomotion is a key capability for robots navigating human spaces. These robots are usually underactuated and have non-smooth and nonlinear dynamics, posing a challenging planning and control problem that has been the focus of significant work over the last several decades. Recent methods have achieved impressive results by reasoning about the full nonlinear dynamics of the robot and the corresponding non-convex optimal control problem, either online using nonlinear model-predictive control (NMPC) or offline using learning-based methods such as reinforcement learning (RL). However, these methods are computationally expensive, difficult to tune, and hard to analyze, leading to an open question: what level of reasoning about these complex nonlinear dynamics is required for locomotion? In this paper we demonstrate that basic locomotion tasks are achievable using a single time-invariant linear dynamics model and convex model-predictive control (MPC).

Some of the earliest legged locomotion work hinted that these challenging systems have significant underlying structure. By leveraging simple models such as the spring-loaded inverted pendulum (SLIP) \cite{blickhan_spring-mass_1989} along with the zero-moment point \cite{vukobratovic_stability_1972}\cite{vukobratovic_zero-moment_2004} and support polygon, researchers were able to construct some of the first successful locomotion controllers for walking and running. These approaches resulted in simple and often closed-form control laws that lent themselves to analysis of their stability and robustness properties \cite{ernst_spring-legged_2009}\cite{buhler_analysis_1988}\cite{spong_further_2003}, and yielded surprising results, such as a blind open-loop controller for stable running over rough terrain \cite{altendorfer_stability_2004}. While critical to developing our understanding of locomotion fundamentals, these methods have been challenging to generalize, and have limited ability to reason about leg inertias, constraints, and disturbances.

To address these limitations, researchers have turned to optimization-based methods. One widely used approach is MPC, which formulates and solves a receding-horizon optimal control problem given dynamics, constraints, and a cost function. These problems have classically been solved using second-order (e.g. Newton-type) methods, \cite{bledt_policy-regularized_2017}\cite{koenemann_whole-body_2015}\cite{neunert_whole-body_2018} though more recently sampling-based methods \cite{alvarez-padilla_real-time_2024} have also achieved promising results. MPC approaches have demonstrated agile and robust behaviors, perhaps most famously demonstrated by Boston Dynamics \cite{robotics_today_recent_2020}\cite{boston_dynamics_atlas_2021}. However, a primary challenge of MPC approaches has been solving large optimization problems fast enough online.

A common approach to solve MPC problems online is sequential quadratic programming (SQP). In SQP, the dynamics and constraints are linearized along the previous solution iterate and a quadratic approximation of the cost function is used to form a quadratic program (QP) \cite{nocedal_numerical_2006}. The solution to this subproblem is then used to compute a new iterate. Computing a solution to these QPs involves solving a potentially large linear system, which is computationally expensive and challenging to parallelize \cite{schubiger_gpu_2020}\cite{bishop_relu-qp_2024}. In addition, while each subproblem may be convex, the general nonlinear MPC problem is often non-convex, with no convergence guarantees. Significant domain knowledge and systems engineering is required to achieve good performance in practice \cite{camacho_nonlinear_2007}.

Given that constructing and solving a QP at each iteration is a significant bottleneck, a natural question to ask is: how much information can we leverage from each subproblem? For example, when running MPC online, the subproblems often don't change significantly between solves if the dynamics are much slower than the control rate. The resulting QP also has an equivalent closed-form linear feedback policy if the active constraint set doesn't change. Some nonlinear MPC approaches have leveraged these facts to reduce the effect of solver delay \cite{grandia_feedback_2019}\cite{li_cafe-mpc_2025}. In addition, many works have shown the power of LQR controllers \cite{tedrake_closed-form_2015} and linearized dynamics models for complex problems like airplane perching \cite{moore_control_2012}. Beyond just the QP setting, linear feedback policies have been shown to be surprisingly capable. For example, \cite{rahme_linear_2021} \cite{krishna_linear_2021} showed that learned linear policies can stabilize gait generators for walking on rough terrain.

In this work, we explore the efficacy of linear models by probing the capability of an MPC controller using a single time-invariant linearized whole-body dynamics model of a legged robot. 
We hypothesize that this convex MPC formulation is sufficient for many practical locomotion tasks that do not require extreme joint angles or body orientations.
Our formulation is: 
\begin{itemize}
    \item \textbf{Convex}, yielding a unique solution in polynomial time;
    \item \textbf{Linear}, with no online nonlinear dynamics or kinematics evaluations;
    \item \textbf{Matrix-inverse-free}, requiring only matrix-vector products and clamp operations online.
\end{itemize}

First, we demonstrate our controller on a range of tasks on a Unitree Go2 quadruped, showing that it is able to recover from perturbations and walk to a goal position with up to 90 degrees of yaw error when given only a walking-in-place reference with no additional higher-level planner. We also test stepping onto a 24 cm box and show the controller is able to handle the change in pitch and foot height.

Second, we demonstrate our controller on a hydraulic humanoid, a more challenging system due to reduced control authority, patch contacts, large leg inertias, and a significant sim-to-real gap. We show that it can stably walk in-place and is able to walk forward with a speed of 0.17 m/s and a stride length of 0.3 m, with a swing phase of 0.6 s and a double support phase of 0.3 s, which cannot be performed quasi-statically.

\begin{figure*}[!b]
    \begin{align}
    \min_{x_{1:N}, u_{1:N-1}, \lambda_{1:N-1}} \sum_{k=1}^{N-1}& 
    % \left(\frac{1}{2} x_k^TQx_k + \frac{1}{2} u_k^TRu_k - x_k^TQ\textcolor{Maroon}{\hat{x}_k}  
     % - u_k^TR\textcolor{Maroon}{\hat{u}_k} \right)  +
    % \frac{1}{2} x_N^TQ_fx_N - x_N^TQ_f\textcolor{Maroon}{\hat{x}_N}
    \frac{1}{2}\left[(x_k-\textcolor{Maroon}{\hat{x}_{k}})^TQ(x_k-\textcolor{Maroon}{\hat{x}_{k}}) +  
    (u_k - \textcolor{Maroon}{\hat{u}_{k}})^TR(u_k - \textcolor{Maroon}{\hat{u}_{k}})\right] + 
    \frac{1}{2}(x_{N}-\textcolor{Maroon}{\hat{x}_{N}})^TQ_f(x_{N}-\textcolor{Maroon}{\hat{x}_{N}}) 
     \label{mpc_formulation} \tag{11}\\
    \text{s.t. }\quad\quad &x_1 = \textcolor{Maroon}{x_{i.c.}} \tag{initial state} \nonumber\\
    &A^+x_{k+1} + Ax_k + Bu_k = \textcolor{Maroon}{d_k} \tag{dynamics} \nonumber\\
    &\textcolor{Maroon}{l_c} \leq Jx_{k+1} \leq \textcolor{Maroon}{h_c} \tag{contact mode constraints} \nonumber \\
    &\textcolor{Maroon}{l_f} \leq u_{k} \leq \textcolor{Maroon}{h_f} \tag{force and torque constraints}\nonumber
    \end{align}
    \captionsetup{labelformat=empty}
    \label{formulation}   
\end{figure*}

\section{Background}

\subsection{Dynamics Model}
We model each legged robot using a floating base, where the configuration $q$ consists of the base position in world coordinates, a body-to-world quaternion, and the joint angles. The velocity vector $v$ consists of the linear and angular velocities of the base in the body frame and the joint velocities. We can relate $\dot{q}$ to $v$ using the following velocity kinematics, where $E(q)$ contains the attitude Jacobian for the quaternion \cite{jackson_planning_2021}.
\begin{align}
    \dot{q} = E(q)v \label{eq:vel_kinematics}.
\end{align}
The continuous-time manipulator equations for a robot with external forces is
\begin{align}
    M(q)\dot{v} + C(q, v) = B\tau + J(q)^T\lambda, \label{eq:manip_eqs}
\end{align}
where $M(q)$ is the mass matrix, $C(q, v)$ is the dynamics bias, $B$ is the input Jacobian mapping actuator torques to generalized forces and $\tau$ are motor torques. $\lambda \in \mathbb{R}^{3N_c}$ is a vector of contact forces in the world frame $(x, y, z)$ for each contact point where $N_c$ is the number of contact points, and $J(q)$ is a Jacobian that maps the external contact forces to generalized forces, which is described in more detail in Section \ref{contact}. These dynamics are put in state-space form and discretized using an integrator such as backward Euler or Runge-Kutta. 

\subsection{Contact Modeling}
Contact dynamics are classically modeled as complementarity constraints, which encode that either the distance between the objects in contact or the contact force is zero. Adding that both the distance and contact force must be non-negative results in the following three constraints.
\begin{align}
    \phi(q) \geq 0, \\
    n \geq 0, \\
    \phi(q)n = 0, \label{eq:complementarity}
\end{align}
where $\phi(q)$ is the signed distance as a function of configuration and $n$ is the normal force.

Solving the general set of constraints above results in a problem that is non-convex and can be ill-conditioned or have singular Jacobians. Instead, we adopt the hybrid approach to model these constraints in our MPC problem: We pre-specify whether objects are in contact or not (referred to as the contact mode) at each time step. We can then encode contact directly through equality and inequality constraints:
\begin{align}
    &\phi(q) \geq 0 \text{,} \quad n = 0 \quad \tag{no contact} \\
    &\text{or} \nonumber\\
    &\phi(q) = 0 \text{,} \quad n \geq 0 \quad \tag{contact} 
\end{align}
By dropping the complementarity constraints and fixing the contact modes, we can convexify the constraints. However, the controller can no longer plan over the ordering of contact events.

\subsection{Trajectory Optimization and Nonlinear MPC}
Trajectory optimization and nonlinear MPC problems can be formulated as:

\begin{align}
    \min_{x_{1:N}, u_{1:N-1}} \sum_{k=1}^{N-1} &\ell_k(x_k, u_k) + \ell_f(x_n) \label{eq:gen_traj_opt} \\
    \text{s.t.}\quad\quad &x_1 = x_{i.c.} \nonumber\\
    &f(x_k, u_k, x_{k+1}) = 0  &\forall k \in 1,\dots,N-1 \nonumber\\
    &x_k \in \mathcal{X}, u_k  \in \mathcal{U} &\forall k \in 1,\dots,N-1 \nonumber
\end{align}
In this problem $x_k = [q_k; v_k]$ is the state at time $k$, $u_k$ is the control, $\ell_k$ is the stage cost, $\ell_f$ is the terminal cost, $f$ is a discrete dynamics constraint, and $\mathcal{X}$ and $\mathcal{U}$ are sets representing general state and control constraints. These problems are typically nonlinear and nonconvex, with no guarantee of finding a globally optimal solution.

\section{Controller Formulation}

In this section, we describe the specific dynamics, constraints, and costs of our linear time-invariant MPC formulation, which results in a quadratic program that is solved online.

\subsection{Low-Level Control}

Many robots, including the Unitree Go2, include low-level PD controllers in their motors that operate with much lower latency than is possible through higher-level torque-control interfaces. While our methodology is capable of fast whole-body torque control, we find that the additional latency introduced by the Go2 torque-control interface limits performance. Therefore, we modify the dynamics \eqref{eq:manip_eqs} to include these low-level PD controller dynamics,
\begin{align}
    &M\dot{v} + C + K_pq + K_d\dot{q} = B\tau ,\\
    &\tau = \tau_{ff} + B^T(K_pq_d + K_d\dot{q}_d) ,\nonumber
\end{align}
where $K_p$ and $K_d$ are diagonal gain matrices and $\tau_{ff}$ is a feed-forward torque command. 

\subsection{Dynamics Linearization}

When linearizing the floating-base state, we use an axis-angle rotation parameterization and define the state error $\Delta x = e(x, x_0)$, where we use quaternion operations to compute the attitude error and vector subtraction otherwise. We combine Eq. \ref{eq:vel_kinematics} and Eq. \ref{eq:manip_eqs} into a continuous-time state-space model and linearize them about $x_0, u_0, \lambda_0$ and then apply backward-Euler integration to construct the following affine linear dynamics model.
\begin{align}
    A^+\Delta x_{k+1} +A\Delta x_k& + B \Delta u_k = d_k \label{eq:lin_dynamics}\\
    \text{where} \; \Delta u_k &= [\Delta \tau_k; \; \Delta\lambda_k] \nonumber
\end{align}

A problem we leave to future work is the best choice of linearization point. Intuitively, we desire one that minimizes the prediction error over the operating space of interest. For our locomotion tasks, we chose to use a standing pose and the corresponding controls and contact forces for $x_0$, $u_0$, and $\lambda_0$. 

\subsection{Locomotion Contact Constraints} \label{contact}

Given a single contact point, let $c(q) :\mathbb{R}^{nq}\rightarrow\mathbb{R}^{3}$ map the configuration of the robot to the world coordinates of the contact point and let $J(q) = \frac{\partial c}{\partial q}$. Our constraint restricts the x and y positions at adjacent knot points to be equal when in contact, letting the MPC controller choose where to place the contact. Assuming that the world frame has the contact height at z = 0, we have the following constraint for each contact point.
\begin{align}
    c(q_{k+1}) - \text{diag}(\begin{bmatrix} 1 & 1 & 0\end{bmatrix}) c(q_k) = 0
\end{align}

The resulting linearized constraint is then as follows.
\begin{align}
    J\Delta q_{k+1} - \text{diag}(\begin{bmatrix} 1 & 1 & 0\end{bmatrix})J\Delta q_k = 0 .
\end{align}
When not in contact, the constraint bounds are set to a lower bound of $(-\infty, -\infty, 0)$ and an upper bound of $(\infty, \infty, \infty)$.

\subsection{Reference Tracking}
We provide whole-body gait references of the state, control, and contact forces for the controller to track, which we represent as $\hat x$ and $\hat u$. Since we have a single linearization point, the MPC dynamics plan over $\Delta x = e(x, x_0)$ and $\Delta u = u - u_0$. Similarly, we define the reference with respect to the linearization point as $\Delta \hat x$ and $\Delta \hat u$. For the rest of this section we'll drop the $\Delta$ for brevity and assume $x$, $u$, $\hat x$ and $\hat u$ are with respect to $x_0$ and $u_0$.

\subsection {Linear Time-Invariant MPC}
The final formulation for the linear time-invariant MPC controller is shown in \eqref{mpc_formulation}. It consists of a quadratic tracking cost, an initial condition constraint, and three constraints at each time step: a dynamics constraint, a contact mode constraint, and a friction-cone force constraint. This problem is a quadratic program and can be written in the following general form,
\stepcounter{equation}
\begin{align}
    \min_z &\frac{1}{2}z^THz + g^Tz \\
    \text{s.t.} \;&l \leq Dx \leq h,
\end{align}

since the constraint Jacobian $D$ and cost Hessian $H$ are fixed. This formulation allows certain QP solvers, such as ReLU-QP and OSQP, to pre-factorize or pre-invert the KKT system offline (which is a $\mathcal{O}(n^3)$ operation) and then perform only back-substitutions or matrix-vector products online to solve the QP (which are $\mathcal{O}(n^2)$ operations). The reference and contact modes can be changed by only modifying the cost gradient $g$ and/or constraint bounds $l$ and $h$, which are shown in dark red in \eqref{mpc_formulation}.

\begin{figure}[t]
    \centering
    \includegraphics[width=\linewidth]{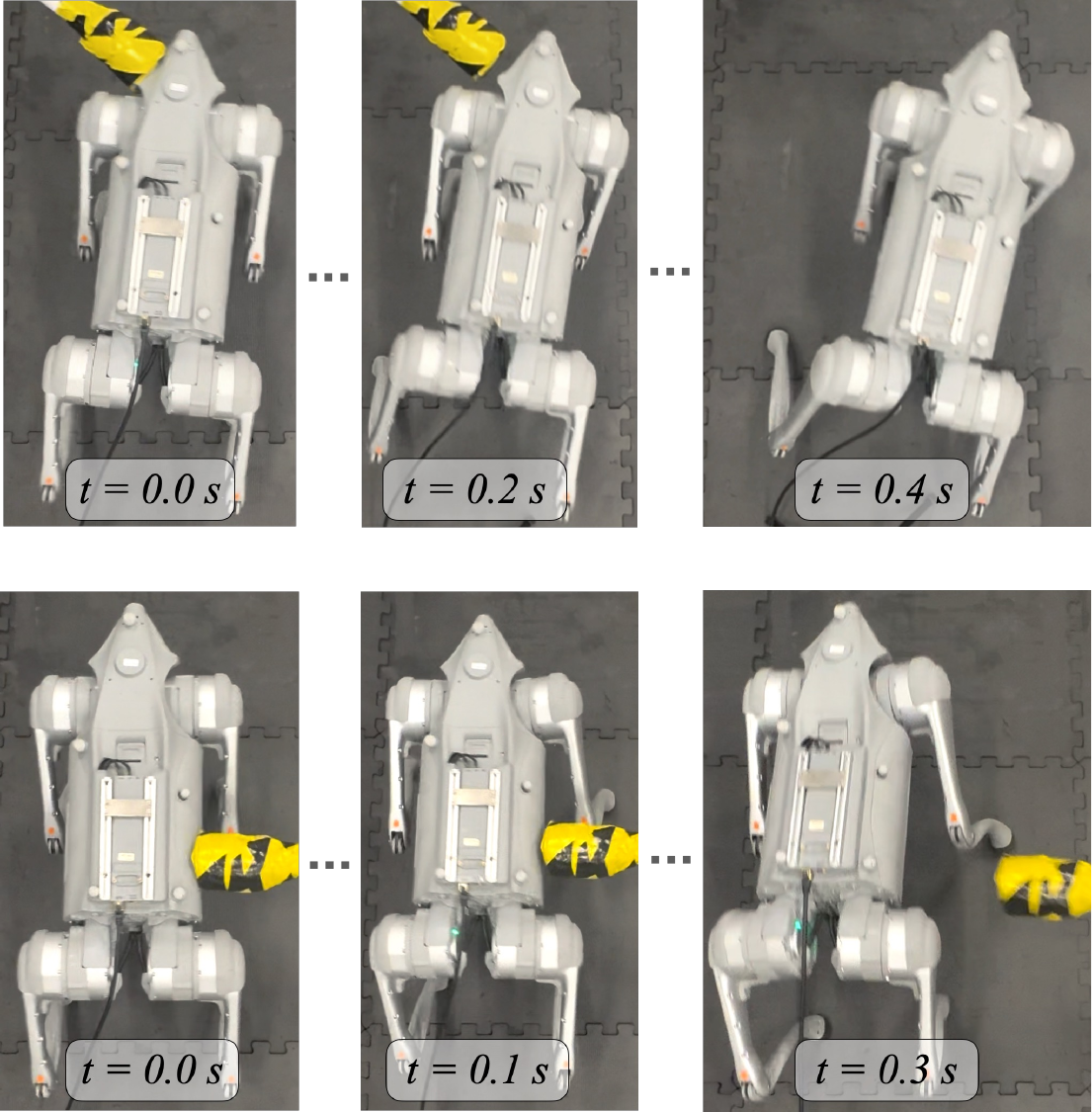}
    \caption{Top-down view of two perturbations applied to the Unitree Go2 while stepping in place. The robot maintains contact with the ground without slipping and return to its original location.}
    \label{fig:home_inplace}
    \vspace{-5mm}
\end{figure}

\section{Experiments}

We tested our MPC formulation on two systems: a quadruped and a humanoid.  Feasible reference motions were planned offline using dynamics from Pinocchio and the trajectory optimization library Aligator \cite{jallet_aligator_nodate}, and the QP was solved online using OSQP \cite{stellato_osqp_2020}. Videos and code can be found on our project website \url{linearwalking.github.io}.

\subsection{Quadruped Experiments}
For our quadruped experiments, we used the Unitree Go2, which has 12 joints (three per leg). It is relatively lightweight, with a total mass of 15.7 kg. While each leg weighs about 2 kg, the mass is centered near the hip, resulting in low angular inertias. We modeled each foot as a single contact point and used constraints on both position and velocity, resulting in 24 constraints and 24 contact forces. The resulting linearized dynamics model has 36 states, 12 actuator torques, and 24 contact forces. 
 
We used a planning horizon of 0.2 seconds with a timestep of 0.01 seconds. The closed-loop control rate was approximately 500 Hz. Experiments were run on a workstation computer equipped with an Intel i9-12900KS CPU and 64 GB of RAM. We performed three experiments: disturbance rejection, goal navigation, and stepping onto a box. For the first two, we used a walking-in-place reference generated with Aligator. No additional online high-level planner or contact heuristics were used during the experiments --- all behaviors came purely from the MPC controller.

\begin{figure}[t]
    \centering
    \includegraphics[width=\linewidth]{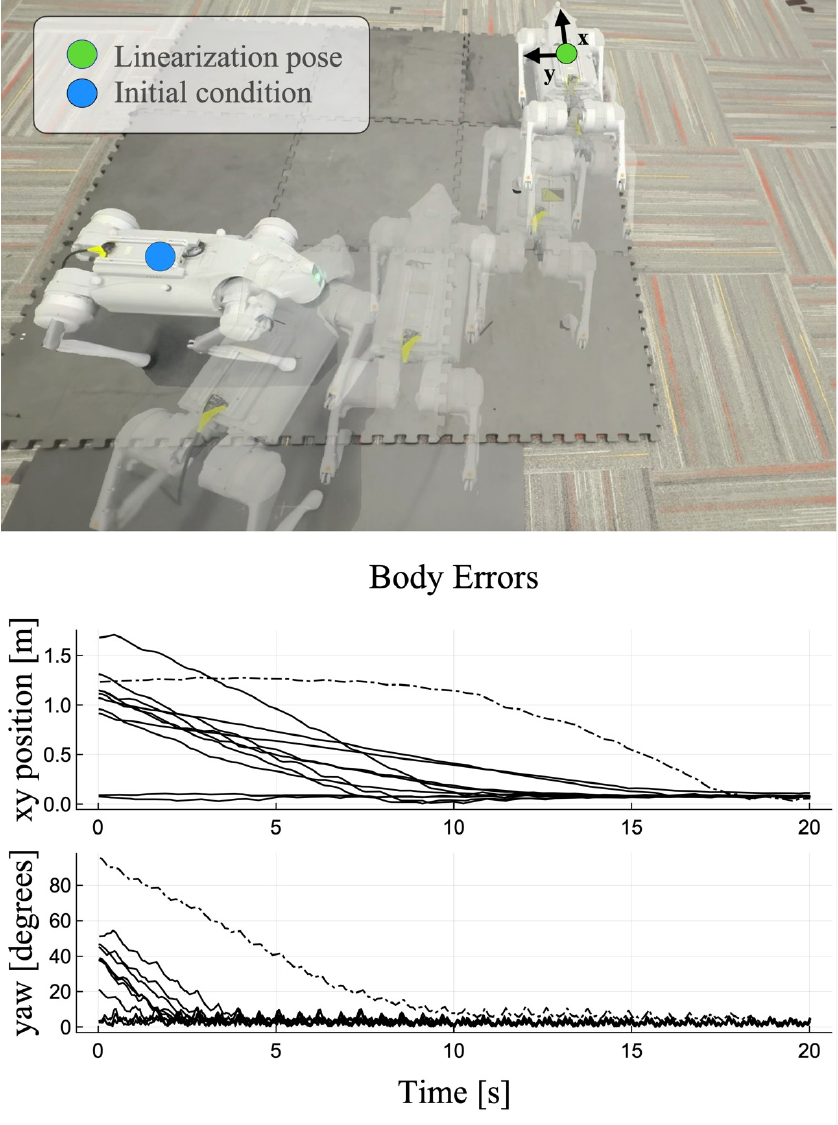}
    \caption{A Unitree Go2 robot starts from 10 initial conditions with yaw from 5 to 90° and displacements from 0.1 to 1.7 m. Transparent overlays show keyframes starting with 90° yaw and 1.3 m displacement. Our controller is able to return to the initial position without assistance from a footstep planner.}
    \label{fig:home_walk}
    \vspace{-5mm}
\end{figure}

\subsubsection{Disturbance Rejection}
We tested disturbance rejection by pushing the robot from the front and side while it stepped in place. Fig. \ref{fig:home_inplace} shows selected frames from the experiments demonstrating that the controller is capable of recovering from moderate pushes and exhibits footstep recovery behaviors despite relying on a single linearization. However, we found that large or sustained perturbations resulted in the controller failing due to an infeasible QP where the contact constraints were not satisfiable. This problem could be alleviated with a footstep planner.

\subsubsection{Goal Navigation}
Using an open-loop walking-in-place reference, we demonstrate that the linear controller is able to plan footsteps to navigate to a goal position despite large changes in orientation. The top of Fig. \ref{fig:home_walk} shows the quadruped starting away from the origin with a yaw error of 90 degrees and walking to the origin over 18 seconds, driving the body position and orientation errors to zero. %We clamp the initial state error online to prevent the controller from trying to take aggressively large steps.
To further validate our controller, we initialized the quadruped from 10 different poses that span yaw errors from five to 90 degrees and 0.1 m to 1.7 m body displacements from the linearization pose. The bottom of Fig.~\ref{fig:home_walk} illustrates the convergence of both the position and orientation errors to zero across all trials, where the dash-dot trajectories correspond with the frames shown in the top figure.

\begin{figure}[t]
    \centering
    \includegraphics[width=\linewidth]{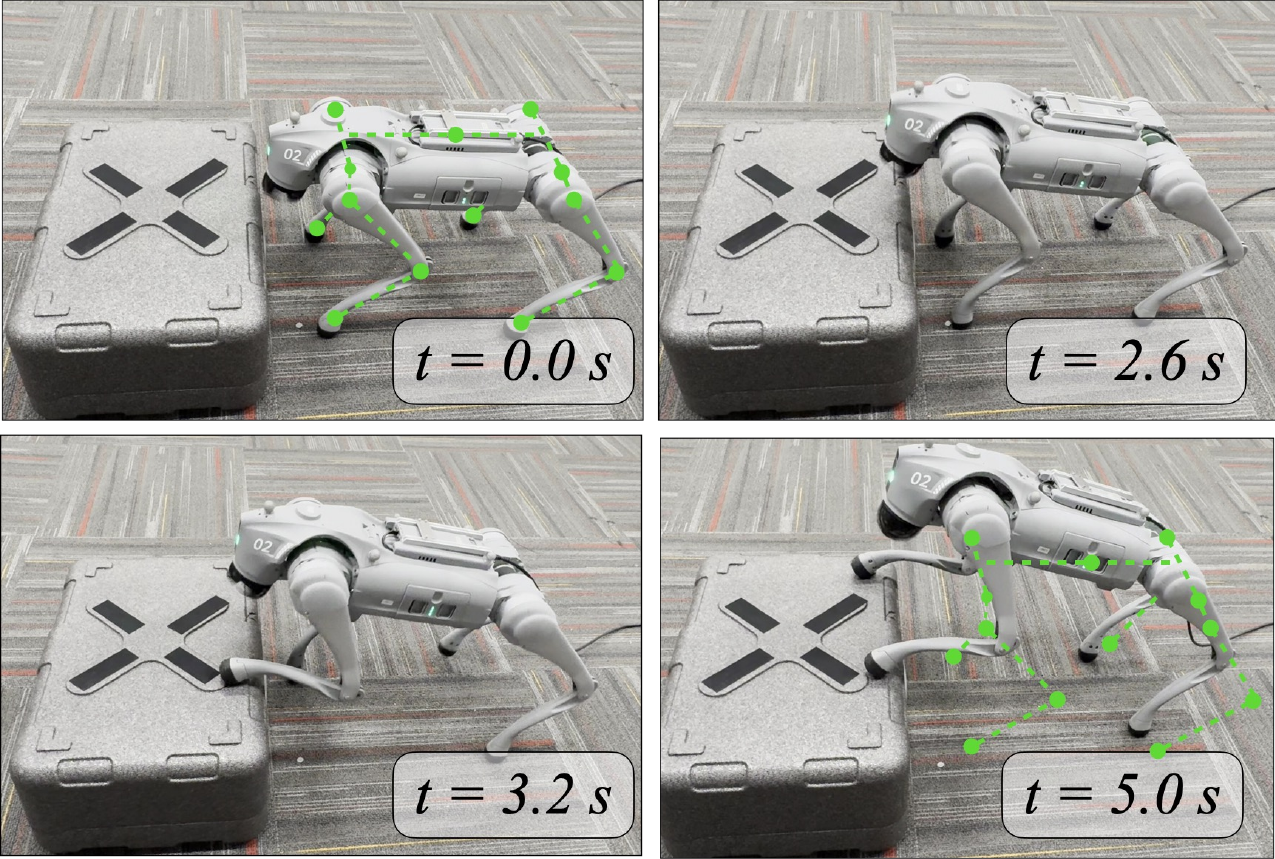}
    \caption{Keyframes of a Unitree Go2 robot stepping onto a 24 cm box. The robot tracks a trajectory with up to 60 degrees joint deviation and 10 degrees pitch deviation from the linearization pose, with the green dashed skeleton showing the linearization pose.}
    \label{fig:box}
    \vspace{-5mm}
\end{figure}

\subsubsection{Stepping onto a Box}
We demonstrate that the controller is able to handle large changes in pitch and footstep height. Fig. \ref{fig:box} shows four keyframes from the robot stepping onto a box with a height of 24 cm, resulting in maximum deviations of 10 degrees in the body pitch, 11 degrees in the hip joints, 29 degrees in the thigh joints, and 60 degrees in the calf joints from the linearization pose. Handling a different contact height is simple, and involves updating the contact constraint from $J_z\Delta x = 0$ to $J_z\Delta x = J_z \Delta \hat x$ where $\hat x$ is the reference.

\begin{figure*}[t!]
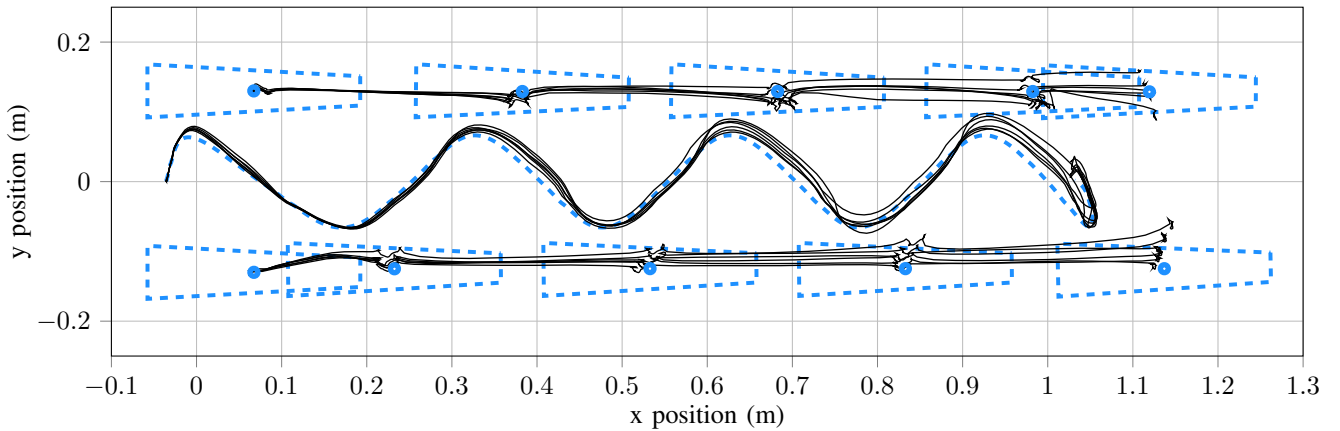

    \centering
    \include{Figures/nadia_tracking}
    \vspace{-10mm}
    \caption{Foot and center-of-mass trajectories across six humanoid walking trials (black) over the desired foot placement positions and center of mass trajectory (blue, dashed). Our controller is able to track the trajectory with minor deviations despite large modeling errors.}
        \label{fig:nadia_tracking}
    \vspace{-4mm}
\end{figure*}
\subsection{Hydraulic Humanoid}

To further test our method, we performed a set of walking experiments on IHMC's Nadia, a hydraulic humanoid. This problem is more challenging for several reasons: Unlike the quadruped, Nadia has hydraulic linear actuators that drive the leg joints through linkage mechanisms. Each joint is approximated as a revolute joint with torque actuators and converted to hydraulic forces by a low-level control stack. While this approximation is exact for most joints, the knee has a four-bar linkage where the revolute joint approximation is inaccurate. Each leg weighs 21 kg, approximately 23\% of the total mass, and has the center of mass close to the knee. In addition, the feet form patch contacts with the ground and the support polygon and capture regions are much smaller \cite{vukobratovic_stability_1972}\cite{vukobratovic_zero-moment_2004}. 

We model each foot with four contact points at its corners. This makes the normal force constraint linear, but the forces are no longer unique and the contact Jacobians are rank deficient. We resolve this issue by adding a small regularizer ($10^{-5}$) to the contact forces constraints. 

For our MPC problem, we chose to focus on control of the lower body, since we had limited access to the hardware. This leaves 12 revolute joints, six for each leg, and one yaw joint at the spine. The resulting linearized dynamics model has 38 states, 13 actuator torques, and 24 contact forces. We planned over a 0.2 second horizon with a timestep of 0.01 seconds, and ran 20 QP iterations for each solve. The closed loop control rate was approximately 333 Hz, which was based off the frequency that the robot's built-in control stack runs at. Experiments were run on a workstation computer equipped with an Intel i7-13700KF CPU.

To approximately quantify the speed gain achieved by using a fixed Hessian and constraint Jacobian and avoiding matrix factorizations online, we measured the factorization and backsolve timing for QDLDL, OSQP's linear system solver, on Nadia's KKT system which is a 3,504 x 3,504 matrix that is 99.3\% sparse. QDLDL takes $5.9 \pm0.8 ms$ to factorize the KKT system but only $0.10 \pm 0.02 ms$ to perform a back-solve. This enables our single convex MPC controller to run significantly faster than an NMPC approach, which would additionally need to evaluate nonlinear dynamics, compute Jacobians, and possibly employ additional steps such as a line-search or outer penalty loops.

We demonstrate our MPC controller on hardware for three different reference motions: walking in-place, and walking forward with short (17 cm) and long (30 cm) stride lengths. The parameters for each reference are shown in Table \ref{table:gait_params}. We found the robot was able to stably walk in place, walking for over a minute with no noticeable instabilities. We also ran six trials for both forward walking references consisting of eight steps each with no failures. Figure \ref{fig:nadia_walk} shows each step from one of the trials on hardware.  Figure \ref{fig:nadia_tracking} shows the foot locations and center of mass trajectories in the XY plane for all 6 runs with a stride length of 30 cm, with the reference foot locations and center of mass trajectory shown with dashed lines. The reference trajectory is not quasi-static, as shown by the center of mass staying out of the support polygon of the stance foot for each step. The maximum final tracking error for the center of mass across all six runs was 3.6 cm. 

\begingroup
\begin{table} 
\small
\begin{center}
\begin{tabular}{c c c c} 

 Reference & Stride & Single Support & Double Support \\ 
 \hline\hline
 In-place & NA & 0.8 s & 0.6 s \\

 Forward short & 17 cm & 0.8 s & 0.6 s \\

 Forward long & 30 cm & 0.6 s & 0.3 s \\

\end{tabular}
\caption{Gait parameters for the three humanoid walking behaviors demonstrated on hardware.}
\label{table:gait_params}
\vspace{-10mm}
\end{center}

\end{table}
\endgroup

\section{Discussion and Conclusions}

We have presented a linear time-invariant MPC controller that is convex and avoids both online nonlinear dynamics evaluations and costly matrix factorizations. We demonstrated the resulting controller on a variety of quadruped tasks to show its ability to handle orientation deviations and reason about foot placement. In addition, we showed that our controller can perform walking on a hydraulic humanoid with large leg inertias and complex kinematics. 

\subsection{Limitations}

We have focused on a simple formulation --- avoiding a higher-level planner and nonlinear evaluations --- to isolate the capabilities of our single-linearization formulation. However, this does result in some clear limitations: Without contact re-planning, large disturbances can lead to infeasible QPs and controller failure, which we observed during large-perturbation experiments. We also found that during bipedal walking the controller struggled to keep the robot's feet flat. In addition, the cost Hessian is also fixed in our approach, preventing task-specific cost tuning, for example for swing-leg tracking.

\subsection{Future Work}

There are many potential directions for future work: First, some of the issues noted above could be addressed with a learned or heuristic-based planner that could update contact modes, preventing infeasible QPs. Second, thanks to its low computational cost, our controller could be embedded directly into the training pipeline of an RL policy reasoning about constraints. Third, some nonlinear dynamics evaluations could also be incorporated to increase performance with very minimal additional computational cost. Lastly, switching between multiple linear models online could enable highly dynamic motions with very large joint-angle or body attitude deviations.

\section{Acknowledgments}
Juan Alvarez-Padilla gratefully acknowledges financial support from the Universidad de Guadalajara (grant V/2022/397) and the Fulbright US Student Program.

\printbibliography

\end{document}